\documentclass[10pt, a4paper]{article}
\usepackage{lrec}
\usepackage{multibib}
\newcites{languageresource}{Language Resources}
\usepackage{graphicx}
\usepackage{tabularx}
\usepackage{soul}
\usepackage[table,xcdraw,x11names]{xcolor}
\usepackage{multirow,booktabs,url}

\usepackage{epstopdf}
\usepackage[utf8]{inputenc}
\usepackage[T2A,T1]{fontenc}
\usepackage[english, russian]{babel}
\usepackage{tempora}
\usepackage{extdash}

\addto\captionsrussian{%
\renewcommand{\abstractname}{Abstract}%
}

\usepackage{hyperref}
\usepackage{xstring}

\usepackage{color}

\newcommand{\scell}[2][c]{%
  \begin{tabular}[#1]{@{}c@{}}#2\end{tabular}}

\title{KazNERD: Kazakh Named Entity Recognition Dataset} 

\name{Rustem Yeshpanov, Yerbolat Khassanov, Huseyin Atakan Varol}

\address{Institute of Smart Systems and Artificial Intelligence,\\Nazarbayev University, Nur-Sultan, Kazakhstan \\
 \{rustem.yeshpanov, yerbolat.khassanov, ahvarol\}@nu.edu.kz\\}

\abstract{
We present the development of a dataset for Kazakh named entity recognition.
The dataset was built as there is a clear need for publicly available annotated corpora in Kazakh, as well as annotation guidelines containing straightforward—but rigorous—rules and examples.
The dataset annotation, based on the IOB2 scheme, was carried out on television news text by two native Kazakh speakers under the supervision of the first author.
The resulting dataset contains 112,702 sentences and 136,333 annotations for 25 entity classes.
State-of-the-art machine learning models to automatise Kazakh named entity recognition were also built, with the best-performing model achieving an exact match F$_1$-score of 97.22\% on the test set.
The annotated dataset, guidelines, and codes used to train the models are freely available for download under the CC BY 4.0 licence from \url{https://github.com/IS2AI/KazNERD}.\\\newline
}

\begin{document}

\maketitleabstract

\section{Introduction}
Named entity recognition (NER) refers to a subtask of information extraction aimed at identifying named entities (NEs) in semi- or unstructured text and classifying them into pre-specified types \cite{Nadeau2007}. NEs, in turn, generally refer to (proper) names of persons, organisations, and geographical locations \cite{tjong-kim-sang-2003-introduction}, as well as numerical and temporal expressions, including quantities, monetary units, percentages, dates, or durations \cite{Chinchor1998}. Widely used in natural language processing applications, including automatic text understanding \cite{Cheng2020}, machine translation \cite{Babych2003}, question answering \cite{Molla2006}, and knowledge base development \cite{Etzioni2005} to name a few, NER has been of interest not only to scientific research, but also to business \cite{Schon2019} and defence \cite{Han2020} ever since 1995, when the term was coined \cite{Grishman1996}. 

By virtue of most of the early works in information extraction being launched as part of United States Government initiatives (e.g., ACE, MUC, TIPSTER) \cite{Maynard2003}, a great deal of research in NER concerns English. Nonetheless, an equally large proportion of NER research has been dedicated to different well-resourced languages, such as Spanish, French, German, Japanese, Chinese, Russian (see \newcite{Nadeau2007}, for a detailed overview), as well as to less resourced ones, such as Sindhi~\cite{DBLP:conf/lrec/AliLX20}, Romanian~\cite{DBLP:conf/lrec/DumitrescuA20}, and Icelandic~\cite{ingolfsdottir-etal-2019-towards}.


Likewise low-resourced, the language of interest of this paper—Kazakh—has only latterly appeared on the radar of NER researchers. Underrepresented and lexically underdeveloped because overshadowed by Russian, which was promoted as a lingua franca during the Soviet era \cite{dave2007kazakhstan}, the earliest NER research in this agglutinative Turkic language dates back as recently as 2016. Although there is evidence for annotated corpus 
construction as part of Kazakh NER research \cite{Akhmed-Zaki2020,Tolegen2016}, to our knowledge, neither of the corpora is publicly available. In addition, none of the studies into Kazakh NER appears to have developed annotation guidelines—or at least adapted those existing in other languages—to take into account cases characteristic of the Kazakh language.

Given this relatively nascent stage of Kazakh NER accompanied by the digital underrepresentation of the language and the lack of freely accessible annotated corpora, it is hoped that our research will fill the existing gaps in the field and thus contribute to its further development. Particularly, we built a dataset consisting of 112,702 sentences from television news, of which 86,246 are unique sentences and 26,456 are their various representations. All sentences in the dataset were manually annotated by two native Kazakh-speaking 
linguists, supervised by the first author. This resulted in the largest Kazakh NE annotated corpus. To assist the annotators in making the right choices when presented with expressions potentially matching NEs, annotation guidelines in Kazakh were developed. The guidelines contain rules for annotating 25 NE types, as well as relatable examples of Kazakh NEs.
Finally, we built four state-of-the-art machine learning models to automatise Kazakh NER, with the highest exact match F$_1$\Hyphdash*score reaching 97.22\% on the test set.

The remainder 
of the paper proceeds as follows: \hyperref[sec2]{Section~2} reviews existing research on Kazakh NER. \hyperref[sec3]{Section~3} discusses data collection and preparation, the development of the guidelines and dataset. \hyperref[sec4]{Section~4} provides the annotated dataset specifications, including the description of NEs, as well as the dataset structure and statistics.
\hyperref[sec5]{Section~5} offers the details of the implemented NER models, the experimental setup, and the evaluation criteria and results. 
\hyperref[sec6]{Section~6} discusses the results of the experiment.
\hyperref[sec7]{Section~7} concludes the paper.


\section{Related Work}\label{sec2}


As mentioned earlier, Kazakh is a digitally low-resourced language, with a small number of (annotated) corpora freely available. That said, recently, there have been progressive efforts made to address such underrepresentation. \newcite{khassanov2020crowdsourced} have built a crowdsourced freely accessible Kazakh speech corpus (KSC) containing 332 hours of transcribed audio. 
In another work, \newcite{mussakhojayeva2021kazakhtts} have constructed the first publicly available large-scale Kazakh text-to-speech synthesis dataset consisting of approximately 93 hours of transcribed audio recordings spoken by male and female professional narrators.

While Kazakh speech processing research has been gathering momentum, thanks to the recent development of publicly available datasets, Kazakh NER research can hardly boast of commensurable progress, which appears to be chiefly due to a lack of such resources. One of the earliest studies into Kazakh NER was conducted by \newcite{Sadykova2016}. 
To build a manually-annotated Kazakh NE corpus, two experts were tasked with labelling 1,000 news articles with a set of seven NEs—namely, (1) person, (2) organisation, (3) location, (4) geopolitical entity (GPE), (5) event, (6) award, and (7) tender—using the brat rapid annotation tool (BRAT) \cite{Stenetorp2012a}. Approximately 3,000 NEs are reported to have been tagged, of which 1,084 were persons, 974 locations, and 973 organisations. However, no breakdown of the remaining NEs is provided in the paper, nor is reference made to the metric applied to achieve an inter-annotator agreement (IAA) score of 0.86–0.89 \cite{Artstein2008}. Another criticism is that, while the annotation guidelines are reported to have been developed specifically for the task, there is no mention of how to access them or the resulting annotated corpus.

\newcite{Tolegen2016} created a Kazakh NE corpus, annotated according to the IOB (Inside, Outside, Beginning) scheme, from 2,500 general news media articles. The corpus is reported to consist of 18,054 sentences and 270,306 words. Annotation was performed using a self-developed web-based tool, with two native Kazakh speakers using the MUC-7 NE task definition \cite{Chinchor1998} as a guide. More than 14,000 NEs were labelled in three categories: 4,292 persons, 7,391 locations, and 2,560 organisations. The IAA measured with Fleiss’ kappa ranged from 0.93 to 0.98 \cite{Fleiss1971}. Furthermore, the scholars conducted an extensive analysis of Kazakh morphological and word type features and were the first to apply a statistical model to Kazakh NER based on conditional random fields (CRFs) \cite{Lafferty2001}, achieving an F$_1$-score of 89.81\%.

The same model was used as a baseline in \newcite{Tolegen2020}, 
where the researchers approached the Kazakh NER task by comparing (1) a bidirectional long short-term memory (BiLSTM) model \cite{Hochreiter1997}, (2) BiLSTM with CRF (BiLSTM-CRF), and (3) a tensor layer-based deep neural network (DNN) model. While the performance of the BiLSTM model yielded a result significantly lower than that of the baseline model (78.76\%), the performance of the BiLSTM-CRF model varied depending on whether or not character embedding was used, 86.45\% and 80.28\%, respectively. The DNN model outperformed the other models, producing an F$_1$\Hyphdash*score of 90.49\%. Although the three models were trained on the annotated corpus built in \newcite{Tolegen2016}, neither of the studies provides information on access to it.

In \newcite{Kozhirbayev2020}, an annotated NE corpus comprising 29,629 sentences was constructed in the IOB format, with the names of persons, organisations, and locations tagged along with Other, a category for NEs of interest that presumably fall outside the three said categories. Four methods to address the Kazakh NER task were applied—specifically, (1) the random forest classifier \cite{Ho1995}, (2) the Naïve Bayes classifier \cite{Friedman1997}, (3) CRFs, and (4) a hybrid method of BiLSTM and CRF. The results show that, while the first two methods achieved an F$_1$-score in the range of 81\% to 89\%, the hybrid method was notably outperformed by the CRFs, 88\% versus 99\%, in turn. However, the study included no information on what guidelines were followed to build the corpus, the quantities of NEs in the corpus, and how, if any, annotation accuracy checks were performed.

\newcite{Kuralbayev2020} compared four NER models—(1)~CRFs, (2) LSTM with character embedding, (3) LSTM-CRF, and (4) bidirectional encoder representations from transformers (BERT) \cite{Devlin2019}—to anonymise 40,000 court decisions in Kazakh and Russian. The names of persons, organisations, locations and addresses were tagged using a self-built annotation tool. The scholars note that the BERT model, which was run without fine-tuning, reached an F$_1$-score of 87\%, with the results of the other models peaking at 82\%. Nevertheless, some notes of caution are warranted here, because, although the model is reported to have achieved high accuracy for both Kazakh and Russian, it was trained exclusively on Russian data. Furthermore, surprisingly, no mention is made of the guidelines used or the IAA assessment, considering that the annotation was carried out by over 150 local university students recruited for the task. Nor is it stated how many NEs were anonymised as a result.

The last study on Kazakh NER we discuss in this paper is by \newcite{Akhmed-Zaki2020}, who applied the BiLSTM, CRF, and BERT methods to a dataset collected from Kazakh online news portals. The dataset was manually annotated using the IOB scheme with four NEs—(1)~persons, (2)~locations, (3) organisations, and (4) other. In this study, too, the BERT model performed the best with an F$_1$\Hyphdash*score of 97.99\%, followed by CRF (94.27\%) and BiLSTM (85.31\%). While the study provides clear information on the parameters of the BERT model and formulae for the precision, recall, and F$_1$-scores computed, it is still limited by the lack of clarity on the volume of the data. Although the dataset built is claimed to consist of 7,153 sentences, the scholars explicitly state that it was split into 6,507, 2,531, and 3,015 sentences for training, validation, and test sets, respectively, which is 12,053 sentences in the aggregate. It is also unclear whether the category Other was used for NEs that were not names of persons, locations, and organisations, but were still of interest (see, e.g., \newcite{Kozhirbayev2020}), or whether it simply referred to a category of words that are not annotated as NEs and are labelled as O in the IOB scheme. 
Much like in the previous studies, no reference is made to the annotation guidelines adhered, the annotators and their backgrounds, the measurement of IAA, and the means of accessing the annotated dataset.

\begin{table*}[!hbt]
\centering
\setlength{\tabcolsep}{1.2mm}
\begin{center}
\begin{tabular}{cll}
\toprule
\multicolumn{1}{c}{\textbf{\scell{Representation\\designations}}} & \multicolumn{1}{c}{\textbf{Example sentence}} & \textbf{Count} \\ 
\midrule
AID & \colorbox{PaleTurquoise2}{«Доу Джонс» \colorbox{white}{\textbf{ORG}}} \colorbox{DarkGoldenrod1}{бес бүтін жүзден сексен алты процентке \colorbox{white}{\textbf{PERC}}} құнсызданды. & 86,246 \\[0.15cm]
BID & \colorbox{PaleTurquoise2}{«Доу Джонс» \colorbox{white}{\textbf{ORG}}} \colorbox{DarkGoldenrod1}{5,86 процентке \colorbox{white}{\textbf{PERC}}} құнсызданды. & 23,969 \\[0.15cm]
CID & \colorbox{PaleTurquoise2}{«Доу Джонс» \colorbox{white}{\textbf{ORG}}} \colorbox{DarkGoldenrod1}{5,86\%-ке \colorbox{white}{\textbf{PERC}}} құнсызданды. & 1,340 \\[0.15cm]
DID & \colorbox{PaleTurquoise2}{Dow Jones \colorbox{white}{\textbf{ORG}}} \colorbox{DarkGoldenrod1}{бес бүтін жүзден сексен алты процентке \colorbox{white}{\textbf{PERC}}} құнсызданды. & 809 \\[0.15cm]
EID & \colorbox{PaleTurquoise2}{Dow Jones \colorbox{white}{\textbf{ORG}}} \colorbox{DarkGoldenrod1}{5,86 процентке \colorbox{white}{\textbf{PERC}}} құнсызданды. & 326 \\[0.15cm]
FID & \colorbox{PaleTurquoise2}{Dow Jones \colorbox{white}{\textbf{ORG}}} \colorbox{DarkGoldenrod1}{5,86\%-ке \colorbox{white}{\textbf{PERC}}} құнсызданды. & 12 \\
\midrule
\addlinespace[0.1cm]
\multicolumn{2}{c}{\textbf{Total number of sentences}} & \textbf{112,702} \\
\bottomrule
\end{tabular}
\end{center}
\vspace{-0.4cm}
\caption{Details of sentence representations, including designations, sentence count of each representation variant, and an example sentence translated as `Dow Jones has depreciated by 5.86\%.'\label{tab:repr}}
\vspace{-0.3cm}
\end{table*}

\section{Annotated Corpus Construction}\label{sec3}

\subsection{Source Data}
The source data were obtained from the television news of the Khabar Agency, a major broadcasting network in Kazakhstan. With the agency’s permission, the Kazakh transcribed text accompanying the original news posted on their official website\footnote{\url{www.khabar.kz}} was collected over the second half of 2020. The news included reports on events in local and international politics, economy, sports, religion, and education that did not necessarily occur during the data collection period, as some news items were also extracted from the agency’s archives. The extracted text\footnote{Tokens denoting speech disfluencies and hesitations (parenthesised) and background noise [bracketed] were retained in the transcribed text.} was not screened for inappropriate content on the assumption that this must have been prudently done by the agency’s content policy department. The text was split sentence-wise—with an identifier assigned to each sentence—and inspected for grammatical and spelling errors (cf. \newcite{Tolegen2016}) and homoglyphs. Duplicate sentences and those containing only Russian utterances were removed; sentences with both Kazakh and Russian utterances were retained, as Kazakh-Russian codeswitching is normal practice in Kazakhstan \cite{Pavlenko2008,DBLP:conf/specom/MussakhojayevaK21}. 
Ultimately, the total number of sentences was 86,246.

\subsection{Sentence Representation}
To enable the developed NER models (see \hyperref[sec5]{Section~5}) recognise instances of the same NE 
regardless of their typographic characteristics (e.g., numerals written in words and digits), the following six sentence representation variants were adopted:
\begin{itemize}
    \item[1)] \textbf{AID} — All sentence elements were recorded in the Cyrillic script\footnote{At the time of writing, the Kazakh language is undergoing a gradual transition from the Cyrillic to the Latin script, with the full transition scheduled to take place between 2023 and 2031.}. Arabic and Roman numerals (e.g., 9 → \textit{тоғыз}, IV → \textit{төрт}, etc.), names of organisations, applications, events, and so on, spelt in Latin characters (e.g., \textit{Bank of America} → \textit{Банк оф Америка}, \textit{Telegram} → \textit{Телеграм}, etc.), terms conventionally spelt in Latin characters (e.g., \textit{PhD} → \textit{ПиЭйчДи}, etc.), and special symbols (e.g., \% → \textit{процент} or \textit{пайыз}) were recorded in Cyrillic words. \vspace{-0.2cm}
    \item[2)] \textbf{BID} — Sentences of the AID representation with numerals recorded in digits. \vspace{-0.2cm}
    \item[3)] \textbf{CID} — Sentences of the BID representation with percentages recorded using the \% symbol. \vspace{-0.2cm}
    \item[4)] \textbf{DID} — Sentences of the AID representation with words conventionally spelt in the Latin script recorded in that script. \vspace{-0.2cm}
    \item[5)] \textbf{EID} — Sentences of the DID representation with numerals recorded in digits. \vspace{-0.2cm}
    \item[6)] \textbf{FID} — Sentences of the EID representation with percentages recorded using the \% symbol. \vspace{-0.1cm}
\end{itemize}
The assigned representation designations, 
as well as example sentences with the resulting quantity of each variant in the dataset are summarised in Table~\ref{tab:repr}. 

\subsection{Annotation Scheme}
The IOB2 scheme—also referred to as BIO—was selected for annotation~\cite{DBLP:conf/eacl/SangV99}.
Under this scheme, each token in text receives one of three tags—namely, \textbf{B}, \textbf{I} or \textbf{O}, indicating whether a token is at the \textbf{B}eginning, \textbf{I}nside or \textbf{O}utside of an annotated extent.
It is similar to the IOB scheme except that a \textbf{B} tag is used at the beginning of every NE extent (see Table~\ref{tab:iob2}).

\begin{table}[!h]
    \begin{center}
    \begin{tabularx}{0.7\columnwidth}{ll}
        \toprule
        \textbf{Tokens} & \textbf{IOB2 tags}\\
        \midrule
        Dow             & B-ORGANISATION\\
        Jones           & I-ORGANISATION\\
        5               & B-PERCENTAGE\\
        ,               & I-PERCENTAGE\\
        86              & I-PERCENTAGE\\
        \%-ке           & I-PERCENTAGE\\
        құнсызданды     & O\\
        .               & O\\   
        \bottomrule
    \end{tabularx}
    \vspace{-0.2cm}
    \caption{Example of the IOB2 annotation scheme\label{tab:iob2}}
    \vspace{-0.2cm}
    \end{center}
\end{table}

\subsection{Annotation Guidelines}
Considering that none of the studies on Kazakh NER provided Kazakh annotation guidelines that our study could rely on to embark on the task, we decided to create such a set of rules. First, we studied some of the most referenced annotation guidelines for NER—particularly, \newcite{Chinchor1998}, \newcite{Brunstein2002}, \newcite{RaytheonBBNTechnologies2004}, \newcite{LinguisticDataConsortium2008}, and \newcite{Weischedel2012}. Next, the first author experimentally annotated a random sample of 2,000 sentences to see what NEs could actually be extracted from the data on hand. Twenty-two NEs described in the guidelines studied were found in the sample. The first draft of the annotation guidelines containing the definition of an NE, information on the valid boundaries of NEs, rules for NE classification, and related examples was prepared in Kazakh.

Later, as a result of the annotator training task, it was decided to tag three more NEs whose examples were found in the news reports annotated. The NEs under consideration were NON\_HUMAN, MISCELLANEOUS, and ADAGE. While the first two had been previously mentioned in the existing annotation guidelines for NER, the decision to tag ADAGE rested upon the relatively frequent use of Kazakh proverbs and sayings in the training sentences. Due adjustments were made to the guidelines, with some rules clarified and supported by comprehensible examples.

It is also worth mentioning at this point that the guidelines were iteratively amended as annotation proceeded. This was partly due to subsequent encounters with cases unconsidered while drafting the guidelines and partly as a result of daily discussions of questions posed by the annotators hired for the task. For a complete list of the 25 NEs and their brief descriptions, see Table \ref{tab:NE}. The final annotation guidelines (in Kazakh) are  available for download from our GitHub repository\footnote{\url{https://github.com/IS2AI/KazNERD}\label{ft:github}}.


\subsection{Annotation Workflow}
Two native Kazakh-speaking linguists received training in NER for two weeks under the supervision of the first author. As part of training, 3,500 sentences from the Khabar agency’s official website were annotated, by following the developed guidelines. The annotation was carried out using the Webanno web-based tool~\cite{Yimam2013} (see~\cite {DBLP:journals/bib/NevesS21} for an extensive review of various tools for annotation). The annotators worked independently on the same version of a text file, which was subsequently reviewed by the first author for annotation divergences and inconsistencies. The final version of the file contained text with annotations approved or modified as appropriate by the first author. During the training period, the IAA score, computed by Webanno, reached a Fleiss’ kappa of 0.94.

The annotation process proceeded for six months, with the annotators labelling 1,500 sentences per day and the first author inspecting these once they were marked \textit{complete} on Webanno. During the period, the IAA score was in the range of 0.95 to 0.97 Fleiss’ kappa. Table~\ref{tab:NE} provides the statistics for annotated NEs.

\begin{table*}[!h]
\centering
\small
\setlength\tabcolsep{0.25cm}
\begin{center}
\begin{tabular}{c|l|l|rr}
\toprule
\multirow{2}{*}{\textbf{No.}} & \multicolumn{1}{c}{\multirow{2}{*}{\textbf{Entity classes}}}& \multicolumn{1}{|c|}{\multirow{2}{*}{\textbf{Definition}}} & \multicolumn{2}{c}{\textbf{Size}}\\
            &                   &                                                                   & \multicolumn{1}{c}{\textbf{\#}}   & \multicolumn{1}{c}{\textbf{\%}} \\
\midrule
1           & (ADA)GE           & Well-known Kazakh proverbs and sayings                            & 196               & 0.14\\[0.025cm]
2           & ART               & Titles of books, songs, television programmes, etc.                 & 2,407             & 1.77 \\[0.025cm]
3           & (CAR)DINAL        & Cardinal numbers, including whole numbers, fractions, and decimals& 29,260            & 21.46 \\[0.025cm]
4           & (CON)TACT         & Addresses, emails, phone numbers, URLs                            & 198               & 0.15 \\[0.025cm]
5           & DATE              & Dates or periods of 24 hours or more                              & 25,446            & 18.66 \\[0.025cm]
6           & (DIS)EASE         & Diseases or medical conditions                                    & 1,272             & 0.93 \\[0.025cm]
7           & (EVE)NT           & Named events and phenomena                                        & 1,658             & 1.22 \\[0.025cm]
8           & (FAC)ILITY        & Names of man-made structures                                      & 2,145             & 1.57 \\[0.025cm]
9           & GPE               & Names of geopolitical entities                                    & 17,543            & 12.87 \\[0.025cm]
10          & (LAN)GUAGE        & Named languages                                                   & 443               & 0.32 \\[0.025cm]
11          & LAW               & Named legal documents                                             & 533               & 0.39 \\[0.025cm]
12          & (LOC)ATION        & Names of geographical locations other than GPEs                   & 2,175             & 1.60 \\[0.025cm]
13          & (MIS)CELLANEOUS   & Entities of interest but hard to assign a proper tag to           & 244               & 0.18 \\[0.025cm]
14          & (MON)EY           & Monetary values                                                   & 4,560             & 3.34 \\[0.025cm]
15          & (NON)\_HUMAN      & Names of pets, animals or non-human creatures                     & 8                 & 0.01 \\[0.025cm]
16          & NORP              & Adjectival forms of GPE and LOCATION; named religions, etc.       & 3,714             & 2.72 \\[0.025cm]
17          & (ORD)INAL         & Ordinal numbers, including adverbials                             & 3,870             & 2.84 \\[0.025cm]
18          & (ORG)ANISATION    & Names of companies, government agencies, etc.                     & 7,587             & 5.57 \\[0.025cm]
19          & (PERC)ENTAGE      & Percentages                                                       & 4,283             & 3.14 \\[0.025cm]
20          & (PER)SON          & Names of persons                                                  & 13,577            & 9.96 \\[0.025cm]
21          & (POS)ITION        & Names of posts and job titles                                     & 6,141             & 4.50 \\[0.025cm]
22          & (PROD)UCT         & Names of products                                                 & 738               & 0.54 \\[0.025cm]
23          & (PROJ)ECT         & Names of projects, policies, plans, etc.                          & 2,111             & 1.55 \\[0.025cm]
24          & (QUA)NTITY        & Length, distance, etc. measurements                               & 3,908             & 2.87 \\[0.025cm]
25          & TIME              & Times of day and time duration less than 24 hours                 & 2,316             & 1.70 \\
\midrule
\multicolumn{3}{c|}{\textbf{Total number of named entities}}                                         & \textbf{136,333}  & \textbf{100 }\\
\bottomrule
\end{tabular}

\vspace{0.1cm}
{\raggedright \textit{\ \ \ \ \ Note.} The parenthesised NE classes will thus be referenced in the tables hereafter.\par}
\end{center}
\vspace{-0.4cm}
\caption{A list of 25 NEs, their short description and statistics\label{tab:NE}}
\vspace{-0.2cm}
\end{table*}

\section{KazNERD Specifications}\label{sec4}

\subsection{Named Entity Descriptions}
The resulting annotated Kazakh NER dataset (hereafter KazNERD) contains 136,333 NEs. As can be seen from Table~\ref{tab:NE}, the top three NEs in 
KazNERD are CARDINAL, DATE, and GPE. None of the previous Kazakh NER studies has labelled the first two classes. The latter class, embracing names of geopolitical entities, has often been conflated with names of geographical locations under the class LOCATION.

Since news reports are normally preceded by (or at least contain) the day or time when a particular event occurred, the frequent use of dates in the dataset was expectable—a total of 25,446 DATE NEs. What is indeed remarkable is the use of numbers in KazNERD. The two classes denoting numbers, CARDINAL (29,260) and ORDINAL (3,870), comprise practically a quarter of the total quantity of NEs in the dataset, a hefty 24.3\%.

Interestingly, in KazNERD, the triad of NEs most commonly labelled in Kazakh NER research—locations (2,175), persons (13,577), and organisations (7,587)—ranks only third through fifth, even when GPE (17,543) is combined with LOCATION. Also worthy of note is the class ADAGE. The class deriving purely from our observations of Kazakh news and hardly fitting the conventional profile of an NE per se (but rather labelled out of scholarly interest) numbered 196 entities in total. This is higher or comparable in size to the classes MISCELLANEOUS, CONTACT, and NON\_HUMAN, previously described as relatively frequent in the NER literature.\\
There are only eight instances of the class NON\_HUMAN, which includes names of creatures other than humans. Such a scarcity of the NEs in the dataset was expected, given that the source data came from television news, which generally reports real-life events. Nevertheless, it was decided to label the NEs as a separate class for consistency with the existing annotation guidelines for NER.\\
As regards MISCELLANEOUS, the class embraces names of school and university subjects, types of computer networks and technologies, livestock breeds, and other entities that we had difficulty in categorising or deemed superfluous to label as separate classes.\\
The remaining NEs in KazNERD have been commonly annotated in the existing NER literature and guidelines. The relatively high ranking of the POSITION class (seventh overall, with 6,141 NEs) can be attributed to the domain of television news, which frequently reports on resolutions and activities of individuals holding official titles and occupational positions. The same applies to news reports on the economy, finance, trade, legal frameworks, business and political objectives, and technology in the country in particular and in the world in general, resulting in NEs annotated for the classes MONEY, PERCENTAGE, QUANTITY, PROJECT, PRODUCT, and LAW, accounting for a total of 11.29\% of all NEs found in KazNERD.\\ 
The names of national and international cultural and political events, as well as the times and venues at which these were held; geographical, ethnic, and religious origins of persons participating in the events among other things; works of art and the languages in which these were produced, are reflected in labelling the classes EVENTS (1,658), NORP (3,714), TIME (2,316), FACILITY (2,145), ART (2,407), and LANGUAGE (443).\\
Lastly, the comparatively frequent instances of the class DISEASE (1,272 NEs) in KazNERD may be explained by two interrelated factors. First, at the time of conducting the present study, the coronavirus disease 2019 (COVID-19) pandemic received massive public attention, which led to the source data often reflecting information on the outbreak of the disease across the country and worldwide. Second, the national media regularly discussed symptoms of various diseases similar to those observed in individuals infected with COVID-19, which resulted in the names of the diseases appearing in the source news reports.

\subsection{Structure and Statistics}
To allow reproducibility of the NER experiment between different research groups, KazNERD was split into three sets: training (80\%), validation (10\%), and test (10\%).
Table~\ref{tab:set_stats} provides statistical information on the number of tokens, sentences, and NEs in the dataset and per set.
An evenly proportional distribution of sentence representations and NEs across the sets was ensured.
We also saw to it that a sentence and its representations were only assigned to the same set.
More detailed information on the numbers of NEs and sentence representations across the three sets can be found in Tables~\ref{tab:sets} and \ref{tab:sets_repr}.

\begin{table}[!b]
\setlength{\tabcolsep}{1.1mm}
\begin{center}
\begin{tabularx}{\columnwidth}{lp{0.3cm}|rrr|r}
    \toprule
                            &   &\multicolumn{1}{c}{\textbf{Train}} & \multicolumn{1}{c}{\textbf{Valid}}    & \multicolumn{1}{c}{\textbf{Test}} & \multicolumn{1}{|c}{\textbf{Total}}\\
    \midrule
    \multirow{2}{*}{\textbf{Sentences}} & \textbf{\#}   & 90,228    & 11,167    & 11,307    & 112,702 \\
                                        & \textbf{\%}   & 80.06     & 9.91      & 10.03     & 100 \\[0.1cm]
    \multirow{2}{*}{\textbf{Tokens}}    & \textbf{\#}   & 1,043,305 & 129,223   & 129,824   & 1,302,352 \\
                                        & \textbf{\%}   & 80.11     & 9.92      & 9.97      & 100 \\[0.1cm]
    \multirow{2}{*}{\textbf{NEs}}       & \textbf{\#}   & 109,342   & 13,483    & 13,508    & 136,333 \\
                                        & \textbf{\%}   & 80.20     & 9.89      & 9.91      & 100 \\
    \bottomrule
\end{tabularx}
\caption{The statistics for the training, validation, and test sets of KazNERD\label{tab:set_stats}}
\end{center}
\end{table}

\begin{figure}[!h]
    \begin{center}
        \includegraphics[width=0.6\linewidth,trim={0.0cm 8.25cm 22.5cm 0.0cm},clip=true]{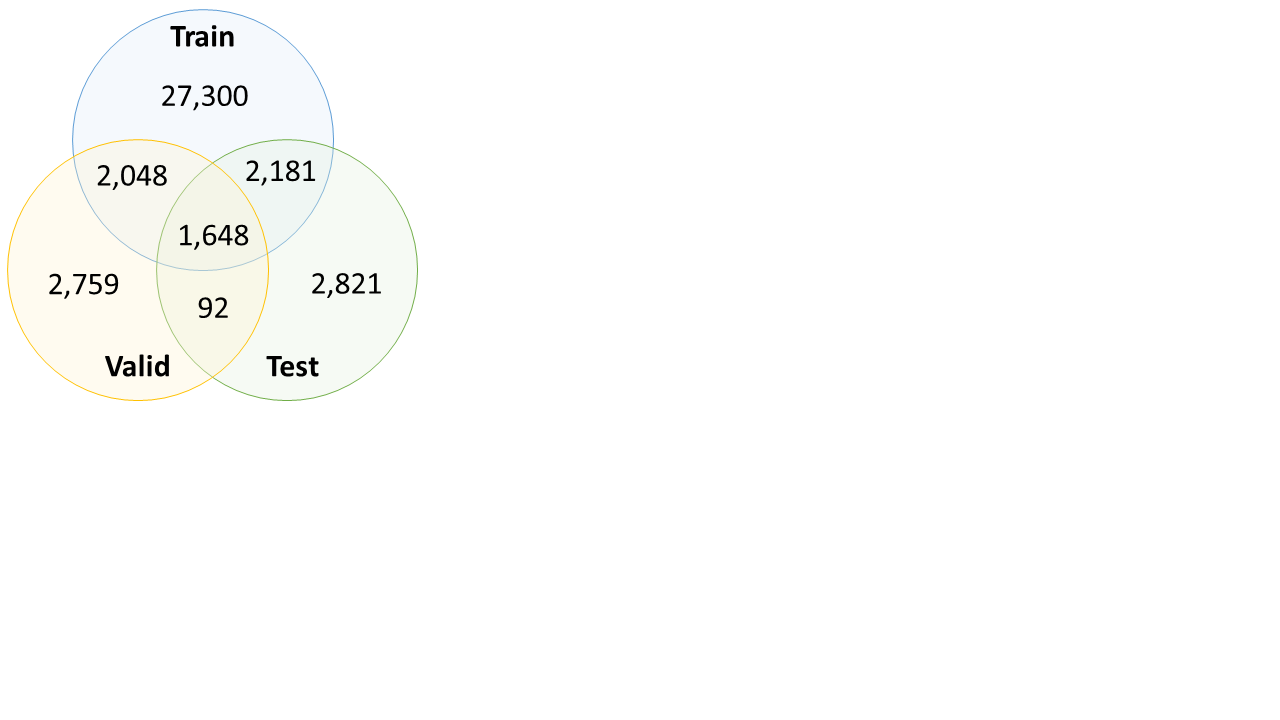}
        \vspace{-0.25cm}
        \caption{A Venn diagram depicting the intersection of unique NEs between the training, validation, and test sets of KazNERD}
        \label{fig:venn}
    \end{center}
\end{figure}

Furthermore, we extracted all unique NEs from KazNERD and computed the intersection between the training, validation, and test sets (see Figure~\ref{fig:venn}).
The total numbers of unique NEs in the training, validation, and test sets are 33,177, 6,547, and 6,742, respectively.
We found that 42\% of the unique NEs in the test set do not appear in the training and validation sets, which confirms its suitability for evaluating the generalisation capability of the NER models.

The three sets are stored in separate files, in the CoNLL-2002 format~\cite{tjong-kim-sang-2002-introduction}—that is, all files contain one token and the corresponding NE tag per line, with blank new lines 
representing sentence boundaries (see Table~\ref{tab:iob2}). Tokens and IOB2 
tags are separated by a single space.
Additionally, we provide variants of the files containing 
identifiers 
heading each sentence, to allow for more nuanced studies requiring representation- and sentence-level detail.
The sentence identifiers are formed by combining representation 
designations (i.e., AID, BID, CID, DID, EID, and FID) with a unique six-digit sentence number
, for example, AID123456.
Sentences with multiple representations have the same six-digit number but different designations, for example, AID111111 and BID111111.

\vspace{-0.3cm}
\begin{table}[!h]
\setlength\tabcolsep{0.13cm}
\begin{center}
\begin{tabularx}{\columnwidth}{l|rr|rr|rr}
\toprule
\multirow{2}{*}{\textbf{\scell{Entity\\classes}}} & \multicolumn{2}{c|}{\textbf{Train}} & \multicolumn{2}{c|}{\textbf{Valid}} & \multicolumn{2}{c}{\textbf{Test}} \\
        & \multicolumn{1}{c}{\textbf{\#}} & \multicolumn{1}{c|}{\textbf{\%}} & \multicolumn{1}{c}{\textbf{\#}} & \multicolumn{1}{c|}{\textbf{\%}} & \multicolumn{1}{c}{\textbf{\#}} & \multicolumn{1}{c}{\textbf{\%}} \\ 
\midrule
ADA     & 159       & 0.15      & 18        & 0.13      & 19        & 0.14 \\[0.025cm]
ART     & 1,953     & 1.79      & 225       & 1.67      & 229       & 1.70 \\[0.025cm]
CAR     & 23,550    & 21.54     & 2,886     & 21.40     & 2,824     & 20.91 \\[0.025cm]
CON     & 160       & 0.15      & 18        & 0.13      & 20        & 0.15 \\[0.025cm]
DATE    & 20,226    & 18.50     & 2,609     & 19.35     & 2,611     & 19.33 \\[0.025cm]
DIS     & 1,031     & 0.94      & 118       & 0.88      & 123       & 0.91 \\[0.025cm]
EVE     & 1,352     & 1.24      & 150       & 1.11      & 156       & 1.15 \\[0.025cm]
FAC     & 1,752     & 1.60      & 195       & 1.45      & 198       & 1.47 \\[0.025cm]
GPE     & 14,108    & 12.90     & 1,693     & 12.56     & 1,742     & 12.90 \\[0.025cm]
LAN     & 352       & 0.32      & 45        & 0.33      & 46        & 0.34 \\[0.025cm]
LAW     & 424       & 0.39      & 54        & 0.40      & 55        & 0.41 \\[0.025cm]
LOC     & 1,759     & 1.61      & 204       & 1.51      & 212       & 1.57 \\[0.025cm]
MIS     & 194       & 0.18      & 24        & 0.18      & 26        & 0.19 \\[0.025cm]
MON     & 3,678     & 3.36      & 441       & 3.27      & 441       & 3.26 \\[0.025cm]
NON     & 6         & 0.01      & 1         & 0.01      & 1         & 0.01 \\[0.025cm]
NORP    & 2,972     & 2.72      & 370       & 2.74      & 372       & 2.75 \\[0.025cm]
ORD     & 3,105     & 2.84      & 379       & 2.81      & 386       & 2.86 \\[0.025cm]
ORG     & 6,093     & 5.57      & 759       & 5.63      & 735       & 5.44 \\[0.025cm]
PERC    & 3,384     & 3.09      & 443       & 3.29      & 456       & 3.38 \\[0.025cm]
PER     & 10,893    & 9.96      & 1,352     & 10.03     & 1,332     & 9.86 \\[0.025cm]
POS     & 4,937     & 4.52      & 601       & 4.46      & 603       & 4.46 \\[0.025cm]
PROD    & 592       & 0.54      & 73        & 0.54      & 73        & 0.54 \\[0.025cm]
PROJ    & 1,694     & 1.55      & 206       & 1.53      & 211       & 1.56 \\[0.025cm]
QUA     & 3,094     & 2.83      & 407       & 3.02      & 407       & 3.01 \\[0.025cm]
TIME    & 1,874     & 1.71      & 212       & 1.57      & 230       & 1.70 \\
\midrule
\textbf{Total}  & \textbf{109,342}  & \textbf{100}      & \textbf{13,483}   & \textbf{100}   & \textbf{13,508}   & \textbf{100} \\
\bottomrule
\end{tabularx}
\caption{The distribution of NEs across the training, validation, and test sets of KazNERD\label{tab:sets}}
\end{center}
\end{table}

\begin{table}[!h]
\setlength\tabcolsep{0.165cm}
\begin{center}
\begin{tabularx}{\columnwidth}{c|rr|rr|rr}
\toprule
\multirow{2}{*}{\textbf{Rep}}    & \multicolumn{2}{c|}{\textbf{Train}}& \multicolumn{2}{c|}{\textbf{Valid}}& \multicolumn{2}{c}{\textbf{Test}} \\
&\multicolumn{1}{c}{\textbf{\#}}&\multicolumn{1}{c|}{\textbf{\%}}&\multicolumn{1}{c}{\textbf{\#}}&\multicolumn{1}{c|}{\textbf{\%}}&\multicolumn{1}{c}{\textbf{\#}}&\multicolumn{1}{c}{\textbf{\%}}\\
    \midrule
    AID             & 69,017            & 76.49             & 8,549             & 76.56         & 8,680             & 76.77 \\[0.025cm]
    BID             & 19,236            & 21.32             & 2,368             & 21.21         & 2,365             & 20.92 \\[0.025cm]
    CID             & 1,059             & 1.17              & 140               & 1.25          & 141               & 1.25 \\[0.025cm]
    DID             & 644               & 0.71              & 81                & 0.73          & 84                & 0.74 \\[0.025cm]
    EID             & 263               & 0.29              & 28                & 0.25          & 35                & 0.31 \\[0.025cm]
    FID             & 9                 & 0.01              & 1                 & 0.01          & 2                 & 0.02 \\
    \midrule
    \textbf{Total}  & \textbf{90,228}   & \textbf{100}      & \textbf{11,167}   & \textbf{100}  & \textbf{11,307}   & \textbf{100} \\
    \bottomrule
\end{tabularx}
\vspace{-0.2cm}
\caption{The distribution of sentence representations across the training, validation, and test sets of KazNERD \label{tab:sets_repr}}
\vspace{-0.7cm}
\end{center}
\end{table}

\begin{table*}[t]
    \renewcommand\arraystretch{1}
    \setlength{\tabcolsep}{3mm}
    \begin{center}
    \begin{tabular}{lccc p{0mm} ccc}
    \toprule
    \multirow{2}{*}{\textbf{Models}} & \multicolumn{3}{c}{\textbf{Valid}} & & \multicolumn{3}{c}{\textbf{Test}} \\\cline{2-4} \cline{6-8}
                            & \textbf{Precision}    & \textbf{Recall}   & \textbf{F$_{1}$-score}&   & \textbf{Precision}    & \textbf{Recall}   & \textbf{F$_{1}$-score} \\
    \midrule
    CRF                     & 93.62                 & 91.93             & 92.77                 &   & 93.20                 & 91.63             & 92.41 \\
    BiLSTM-CNN-CRF          & 94.51                 & 93.72             & 94.11                 &   & 93.84                 & 93.18             & 93.51 \\
    BERT                    & 96.30                 & 96.07             & 96.19                 &   & 96.14                 & 96.34             & 96.24 \\
    XLM-RoBERTa             & \textbf{97.20}        & \textbf{97.18}    & \textbf{97.19}        &   & \textbf{97.09}        & \textbf{97.35}    & \textbf{97.22} \\                            
    \bottomrule
    \end{tabular}
    \vspace{-0.0cm}
    \caption{Experiment results of four NER models on the validation and test sets of KazNERD 
    \label{tab:results}}
    \vspace{-0.5cm}
    \end{center}
\end{table*}

\section{NER Experiment}\label{sec5}

\subsection{NER Methods}
We applied several state-of-the-art machine learning methods to evaluate the KazNERD corpus.
Detailed information on the NER model implementation and feature construction can be found in our GitHub repository\textsuperscript{\ref{ft:github}}.

\textbf{CRF} 
We applied the CRF models implemented by the CRFsuite toolkit~\cite{okazaki2007crfsuite}.
Specifically, we used the features derived from the surface forms of tokens, including target and context token prefixes, suffixes, and shape features.
We note that the CRF models do not incorporate external linguistic resources, such as gazetteers, lookup tables, or word vector features.

\textbf{BiLSTM-CNN-CRF}
We used the PyTorch implementation of a BiLSTM-CNN-CRF model~\cite{DBLP:conf/acl/MaH16}.
The model combines the word embeddings with the character-level representations extracted using the CNN and feeds them into the BiLSTM module with the CRF output layer.
Word embeddings are usually pre-trained on large unlabelled corpora, but, in the present study, we used randomly initialized embeddings.

\textbf{BERT}
A pre-trained BERT model can be readily applied to the NER task, by reinitializing the output layer size to match the NE labels and fine-tuning the model on the NER data.
We used the case-sensitive version of the multilingual BERT model within the Hugging Face Transformers framework~\cite{wolf-etal-2020-transformers}.
The model consists of around 110M parameters and was pre-trained on 104 languages with the largest Wikipedia content, which includes the Kazakh language as well.

\textbf{XLM-RoBERTa}
We also applied the XLM-RoBERTa model~\cite{DBLP:conf/acl/ConneauKGCWGGOZ20}, a multilingual version of RoBERTa~\cite{liu2019roberta}, within the  Hugging Face Transformers framework.
Similar to BERT, it was adapted for the NER task, by reinitializing the output layer and fine-tuning.
The rationale behind choosing the model lies in the fact that it has over five times as many parameters as BERT does (560M) and was pre-trained on CommonCrawl data containing 100 languages, Kazakh included.

\subsection{Experimental Setup}
The four NER models were trained on the training set. The hyperparameters were tuned on the validation set. The final, best-performing, model was evaluated on the test set. 
The deep learning-based models utilised a single V100 GPU on an NVIDIA DGX-2 machine.

The CRF model was run for 550 iterations using the L-BFGS training algorithm, with the \textit{L$_1$} and \textit{L$_2$} regularisation terms set to 0.1 and 0.01, respectively.
The other hyperparameters were left at their default values of the CRFsuite toolkit. 

For the BiLSTM-CNN-CRF model, we used a single BiLSTM layer with 256 hidden units and a CNN layer with 30 filters of size 3.
The word and character embedding sizes were set to 100 and 30, respectively.
We chose an initial learning rate of 0.005 and a batch size of 1,024.
To prevent overfitting, a dropout rate of 0.5 was applied. 
The model was trained for 1,000 epochs using the Adam optimizer and the early stopping criteria based on the validation set, which yielded the highest score on epoch 432.


The BERT model was fine-tuned for 8 epochs, with the initial learning rate set to $5\cdot10^{-5}$ and the weight decay rate set to $10^{-4}$.
We set the batch size to 128 and applied 3,000 warmup steps.
Likewise, the XLM-RoBERTa model was fine-tuned for 10 epochs, with the initial learning rate set to $10^{-5}$ and the weight decay rate set to $10^{-3}$.
We set the batch size to 64 and applied 800 warmup steps.

\subsection{Evaluation Criteria}
We evaluate NER performance in terms of exact match using precision, recall and F$_1$-score~\cite{Nadeau2007} and the standard seqeval script~\cite{seqeval}, requiring that both the type and span of predicted NEs match the gold standard mention.

\subsection{Experiment Results}
Table~\ref{tab:results} presents the performance of the 
NER models on the validation and test sets of KazNERD, measured by micro-averaging \cite{yang2001study}.
The highest results were achieved by XLM-RoBERTa, followed by BERT, BiLSTM-CNN-CRF and CRF.
Specifically, XLM-RoBERTa achieved relative improvements of 1\%, 4\%, and 5\%  over BERT, BiLSTM-CNN-CRF, and CRF, respectively. 
In general, all the NER models performed well, achieving precision, recall, and F$_{1}$-scores of above 90\%, highlighting the utility of our annotated dataset for the Kazakh NER task.
The results of the XLM-RoBERTa model for different NEs are shown in  Table 8 and will be discussed in the following section.

\begin{table}[!h]
    \setlength{\tabcolsep}{3.0mm}
    \begin{center}
    \begin{tabularx}{0.9\columnwidth}{lccc}
        \toprule
        \textbf{\scell{Entity\\Classes}}& \textbf{Precision}& \textbf{Recall}   & \textbf{F$_{1}$-score}    \\
        \midrule
        ADA                             & 83.33             & 52.63             & 64.52 \\[0.025cm]
        ART                             & 97.83             & 98.25             & 98.04 \\[0.025cm]
        CAR                             & 98.48             & 98.90             & 98.69 \\[0.025cm]
        CON                             & 89.47             & 85.00             & 87.18 \\[0.025cm]
        DATE                            & 97.49             & 98.01             & 97.75 \\[0.025cm]
        DIS                             & 90.84             & 96.75             & 93.70 \\[0.025cm]
        EVE                             & 87.27             & 92.31             & 89.72 \\[0.025cm]
        FAC                             & 79.21             & 80.81             & 80.00 \\[0.025cm]
        GPE                             & 98.38             & 97.59             & 97.98 \\[0.025cm]
        LAN                             & 95.74             & 97.83             & 96.77 \\[0.025cm]
        LAW                             & 87.04             & 85.45             & 86.24 \\[0.025cm]
        LOC                             & 91.63             & 87.74             & 89.64 \\[0.025cm]
        MIS                             & 96.15             & 96.15             & 96.15 \\[0.025cm]
        MON                             & 99.77             & 100.00            & 99.89 \\[0.025cm]
        NON                             & 0.00              & 0.00              & 0.00 \\[0.025cm]
        NORP                            & 98.92             & 98.92             & 98.92 \\[0.025cm]
        ORD                             & 97.39             & 96.63             & 97.01 \\[0.025cm]
        ORG                             & 91.84             & 93.47             & 92.65 \\[0.025cm]
        PERC                            & 98.68             & 98.68             & 98.68 \\[0.025cm]
        PER                             & 99.55             & 99.17             & 99.36 \\[0.025cm]
        POS                             & 96.29             & 99.00             & 97.63 \\[0.025cm]
        PROD                            & 88.89             & 87.67             & 88.28 \\[0.025cm]
        PROJ                            & 93.81             & 93.36             & 93.59 \\[0.025cm]
        QUA                             & 97.30             & 97.30             & 97.30 \\[0.025cm]
        TIME                            & 98.69             & 98.26             & 98.47 \\[0.025cm]
        \midrule
        \textbf{\scell{Micro ave.}}  & \textbf{97.09}    & \textbf{97.35}    & \textbf{97.22} \\[0.1cm]
        \textbf{\scell{Macro ave.}}  & \textbf{90.16}    & \textbf{89.20}    & \textbf{89.53} \\
        \bottomrule
    \end{tabularx}
    \vspace{-0.0cm}
    \caption{XLM-RoBERTa performance for different entity classes of the test set\label{tab:err_analysis}}
    \vspace{-0.7cm}
    \end{center}
\end{table}

\section{Discussion}\label{sec6}
The performance of XML-RoBERTa was above 95\% for 14 NE classes and in the range of 85\% to 95\% for eight classes. Only three classes were predicted with an F$_1$-score below 85\%. The model yielded almost perfect results for MONEY (99.89\%) and PERSON (99.36\%). This could be explained by the composition of these classes. The extent of MONEY NEs includes a monetary value and an explicit monetary unit (e.g., 50 dollars). This must have made it easier for the model to recognize the class, for monetary units in KazNERD are not very diverse, with “tenge” (the local currency), “dollar”, and “euro” making frequent appearances. Likewise, in Kazakh, PERSON NEs often appear as a combination of first and last names, with both capitalised and the latter normally ending in \textit{–ов(а)} “-ov(a)”, \textit{-ев(а)} “-ev(a)”, \textit{-ин(а)} “-in(a)”. These features presumably enabled the model to achieve high prediction accuracy for the class.

The low F$_1$-scores for NON\_HUMAN (0\%) and ADAGE (64.52\%) on the test set could be due to the apparent insufficiency of instances of the former in the dataset and the form variability of the latter. Increasing the number of NON\_HUMAN NEs in the training sample, by expanding the dataset to embrace domains where the use of names of non-humans is expected (e.g., science fiction, children’s stories, or animal fantasies) will likely improve the accuracy of the model. As for ADAGE NEs, they are generally easy to recognise in context thanks to their form fixedness (e.g., \textit{No smoke without fire}). Lexical and grammatical variations of proverbs and sayings are possible (e.g., \textit{There is no smoke without fire} or \textit{Where there is smoke, there is fire}), but still unlikely to preclude humans from continuing to identify these: such phrases bear greater psychological and social significance than do other set expressions \cite{Norrick+2015+7+27}. However, this can hardly apply to a machine learning model, which will struggle to decide whether a given expression is a pre-existent variation of a known adage, its nonce restructuring, or not an adage at all, especially if there is inadequacy of data to make inferences from. As mentioned earlier, the class ADAGE was labelled as a result of our scientific curiosity, and further review and investigation as to the worth of this class for the NER task is required.

Since the present study was, to the best of our knowledge, the first to develop a publicly available annotated corpus as well as guidelines in Kazakh for 25 NE classes, it was subject to several challenges. Firstly, although NER generally implies the recognition of proper nouns in text, which are expected to be capitalised given their designation of names of persons, places, organisations and so forth, some Kazakh nouns assigned to certain NE classes in our dataset do not seem to meet this criterion. For example, the NEs \textit{дүйсенбі} “Monday” (DATE), \textit{христиандар} “Christians” (NORP), or \textit{ағылшын тілі} “English” (LANGUAGE) to name a few, are normally lower-cased in Kazakh, unless they appear at the beginning of a sentence. Further studies on Kazakh NER taking such cases into account need to be undertaken.

NE coordination posed another problem. The challenge concerns whether two (or more) coordinated NEs, for example, \textit{Олжас пен Аина Қорғанбек} “Olzhas and Aina Qorganbek” (the names of a husband and a wife followed by their family name; PERSON) or \textit{Байтұрсынов пен Қонаев көшелерінде} “on Baitursynov and Qonayev Streets” (the names of two local streets followed by the word “streets”; FACILITY) ought to be labelled as a single NE or two separate NEs. Although MUC-6 \cite{Grishman1996} originally advocated the separate use of annotations, in KazNERD, it was decided to label coordinated NEs as a single entity in accordance with the recommendations of MUC-7 \cite{Chinchor1998}, promoting joint annotation.

Another similar issue was related to nested entities: for example, should the expression \textit{Қазақстан Президенті} “The President of Kazakhstan” be considered two entities \textit{Қазақстан} (Kazakhstan, GPE) and \textit{Президенті} (President, POSITION) or a single entity \textit{Қазақстан Президенті} (The President of Kazakhstan, POSITION)? Here again, our decision was guided by MUC-7, encouraging the annotation of such expressions as a single NE. Thus, while developing KazNERD, we chose not to decompose compound entities and not to label subentities. However, future research into Kazakh NER should consider these challenges, with the decision as to which of the approaches is more likely to cover the needs of application areas left to the discretion of those concerned.

As regards challenges related to metonymy (i.e., the use of the name of something to refer to that of something else that is closely associated with it, as in \textit{Downing Street} to refer to the British Prime Minister), consistent with the MUC recommendations, KazNERD generally retains the semantics of 
common NEs, unless otherwise specified in the developed annotated guidelines. Thus, in \textit{Абайды тану} “cognising Abai” (the name of a great Kazakh poet), the NE \textit{Абайды} is tagged as PERSON, despite the contextual reference to the person’s literary works (the NE class ART). This should certainly be borne in mind by 
enthusiasts willing to make use of KazNERD.

Similarly, challenges presented by the ambiguity between the classes ORGANISATION and FACILITY may presumably account for the comparatively low F$_1$-score for the latter. In the annotation guidelines, we recommend that, in cases of confusion, preference should be given to ORGANISATION when actions normally characteristic of persons (e.g., \textit{say}, \textit{state}, \textit{report} etc.) are used with names of institutions or if a building houses an institution of the same name, unless explicitly referring to the physical structure alone in a locative manner. Yet, in cases where the distinction is still not clear-cut, such as \textit{Президент ... Ақордада арнайы кеңес өткізді} “President ... held a special meeting in Akorda” (the official workplace of the President of Kazakhstan), we annotated \textit{Ақордада} as ORGANISATION in line with the existing guidelines tagging \textit{White House} or \textit{Kremlin} as ORGANISATION, in spite of the contextual reference to the facility.





\section{Conclusion}\label{sec7}
The present study set out to develop the first publicly available annotated dataset for Kazakh NER. The resulting dataset, KazNERD, contains 112,702 sentences from the television news domain and 136,333 annotations for 25 entity classes. All NEs were labelled using the IOB2 scheme by two native Kazakh speakers under the supervision of the first author, in accordance with the annotation guidelines specially designed in and for the Kazakh language. To automate Kazakh NER, state-of-the-art machine learning models were built, with the best-performing model yielding an exact match F$_1$-score of 97.22\% on the test set. In the future, we aim to focus on developing fine-grained and domain-independent NER models to ensure their external validity. To this end, we intend to train the models on a version of KazNERD supplemented with annotated data from different domains and genres, including transcribed conversations from television and radio shows, podcasts, phone talks, fiction, and senate speeches.

The annotated dataset, guidelines, and codes used in training the models can be freely downloaded under the CC BY 4.0 licence from \url{https://github.com/IS2AI/KazNERD}.

\section{Acknowledgements}\label{sec8}
Our very special thanks go to Aigerim Kabduluakhitova and Aizhan Seipanova—the two annotators, who demonstrated their expertise, diligence, and continued patience throughout the whole process of developing KazNERD.

\section{Bibliographical References}\label{reference}

\bibliographystyle{lrec}
\bibliography{main}


\end{document}